\pdfoutput=1

\documentclass[11pt]{article}

\usepackage{acl}

\usepackage{times}
\usepackage{latexsym}

\usepackage[T1]{fontenc}

\usepackage[utf8]{inputenc}

\usepackage{microtype}

\usepackage{inconsolata}

\usepackage{amsmath, amsfonts, amssymb, amsbsy, mathtools, nccmath, bm}

\usepackage{graphicx}
\usepackage{subcaption}
\usepackage{afterpage}

\usepackage{multirow}
\usepackage{booktabs}
\usepackage{dblfloatfix}

\usepackage{dsfont}

\providecommand{\abs}[1]{\lvert#1\rvert}

\makeatletter
  \newcommand\scripty{\@setfontsize\scripty{6pt}{7}}
\makeatother

\usepackage{hyperref}
\definecolor{darkblue}{rgb}{0.0, 0.0, 0.45}
\hypersetup{colorlinks=true,linkcolor=darkblue,citecolor=darkblue,filecolor=darkblue,urlcolor=darkblue,pagebackref=true,breaklinks=true,bookmarks=true}

\usepackage[multiple]{footmisc}
\makeatletter
\renewcommand\@makefntext[1]%
    {\makebox[4pt][r]{\textsuperscript{\@thefnmark}\,}#1}
\makeatother

\def\model{GRAM}
\providecommand{\modelv}[1]{$M_{\text{#1}}$}
\providecommand{\modelw}[2]{$M_{\text{#1},\text{#2}}$}

\providecommand{\modelvsmall}[1]{$S_{\text{#1}}$}
\providecommand{\modelvtiny}[1]{$T_{\text{#1}}$}
\providecommand{\modelvdeconly}[1]{$DO_{\text{#1}}$}
\providecommand{\modelvencdec}[1]{$ED_{\text{#1}}$}

\usepackage{xcolor}
\providecommand{\ned}[1]{#1}

\usepackage{thanks-nostar}

\title{Adding Multimodal Capabilities to a Text-only Translation Model}

\author{Vipin Vijayan$^{\text{1}}$, Braeden Bowen$^{\text{1}}$, Scott Grigsby$^{\text{1}}$, Timothy Anderson$^{\text{2}}$,
Jeremy Gwinnup$^{\text{2}}$\\
$^{\text{1}}$PAR Government Systems Corporation, $^{\text{2}}$Air Force Research Laboratory\\
\texttt{\small \{vipin\_vijayan, braeden\_bowen, scott\_grigsby\}@partech.com},\\\texttt{\small \{timothy.anderson.20, jeremy.gwinnup.1\}@us.af.mil}}

\begin{document}
\maketitle
\begin{abstract}

While most current work in multimodal machine translation (MMT) uses the Multi30k dataset for training and evaluation, we find that the resulting models overfit to the Multi30k dataset to an extreme degree.
Consequently, these models perform very badly when evaluated against typical text-only testing sets such as the newstest datasets.\\
In order to perform well on both Multi30k and typical text-only datasets, we use a performant text-only machine translation (MT) model as the starting point of our MMT model. We add vision-text adapter layers connected via gating mechanisms to the MT model, and incrementally transform the MT model into an MMT model by 1) pre-training using vision-based masking of the source text and 2) fine-tuning on Multi30k.\\
We achieve a state-of-the-art performance on the Multi30k 2016 en-de test set of 46.5 BLEU4 score and 0.61 CoMMuTE score via this approach while retaining the performance of the original text-only MT model against the newstest dataset.

\end{abstract}

\section{Introduction}
\label{sec:intro}

The task of multimodal machine translation (MMT) is to automatically translate text while using additional modalities (e.g., image, video, audio) to aid in translation.
Prior work has shown that MMT can use contextually relevant images to aid in translation of sentences that contain ambiguities or missing textual information \citep{caglayan_probing_2019, wu_good_2021}. For example, the noun ``bank'' is ambiguous and contextually dependent in English (``financial institution'' or ``river edge'') but unambiguous in French (``\textit{banque}'' or ``\textit{rive}'').
The hypothesis that these ambiguities or missing information can be resolved with contextually relevant images is persuasive.

\ned{Much work} in MMT \citep{yao_multimodal_2020, yin_novel_2020, wu_good_2021, li_valhalla_2022} focus on the Multi30k dataset \citep{elliott_multi30k_2016}, a dataset comprising 30,014 image captions and corresponding translations in different languages.%

However, compared to the domain of text-only translation where MT models are trained using millions of examples, the Multi30k dataset is an extremely small dataset.
Consequently, the MMT models will naturally overfit to the Multi30k dataset and perform poorly against testing sets that text-only translation models are typically evaluated against (Section \ref{sec:experiments}).

Text-only machine translation is a much larger domain than multimodal machine translation and many strong models have been developed in the field \citep{kocmi_wmt_findings_2022}.
Thus, using a pre-trained text-only model as a starting point for MMT is a promising approach to advance the state of MMT.
To demonstrate this, we incrementally transform a text-only MT model into an MMT model, resulting in state-of-the-art performance against the Multi30k dataset while retaining the performance of the pre-trained model against text-only test sets.

We use a pre-trained Transformer-based translation model as our starting point. We evolve this text-only translation model into an MMT model using adapters \citep{houlsby_parameter-efficient_2019} and gating mechanisms such that the model learns how to use visual information while preserving its original translation performance.
We do this by 1) combining a strong pre-trained translation model and a pre-trained vision-language model to create an MMT model, 2) pre-training the MMT model on a dataset of captions augmented with informed visual grounding and machine generated translations along with a dataset collated from a text-only MT dataset, and 3) fine-tuning against the Multi30k dataset. 

Using this model architecture and training process, we achieve high performance against the Multi30k test sets while retaining high performance against text-only testing sets (Table \ref{tab:main}).

\section{Related Works}
\label{sec:related}

\subsection{Adapting pre-trained models for MMT}

\citet{caglayan_cross-lingual_2021} converted a translation language model into a vision-based translation language model by pre-training using Conceptual Captions \citep{sharma_conceptual_2018}, translating English captions to German using a translation model, and fine-tuning using Multi30k. 

\citet{futeral_tackling_2022} also proposed a model that adapts a language model into an MMT model by simultaneously training against the MMT objective using the Multi30k dataset and the visually-conditioned masked language modeling objective using the Conceptual Captions dataset. While they used a visual-conditioned masked language modeling object, we use the much simpler training process of directly optimizing the output using cross-entropy loss. Furthermore, while they randomly choose words for visual grounding, we choose vision-based words selected using an object detection method for our masking.

\subsection{Masking for visual grounding}

Masking words for visual grounding is a common approach employed by such works as \citet{wu_good_2021}, \citet{ive_distilling_2019}, \citet{caglayan_probing_2019}, \citet{wang_efficient_2021}. We cover a subset of these works.

\citet{ive_distilling_2019} masked specific words (ambiguous, inaccurate, and gender-neutral words) in the English source text to force the MMT models to use the visual information to generate target texts. They show that the additional visual context was helpful in text generation. 

\citet{caglayan_probing_2019} performed masking based on color deprivation, whole entity masking, and progressive masking on source texts. However, they found that training based on masking results in performance degradation on the Multi30k testing sets, which indicates that the vision information was not being fully utilized by their models. 

\citet{wang_efficient_2021} performed masking of source text based on Flickr30k-Entities \citep{plummer_flickr30k_2016} that were vision related and used a multi-task object to train their MMT model, where they optimized for object-masking loss in addition to the text generation.

\subsection{Gating mechanism for MMT}

Similar to our work, \citet{wu_good_2021}, \citet{zhang_neural_2020}, \citet{lin_dynamic_2020} and \citet{yin_novel_2020} use a trainable gating mechanism in the context of MMT to control the fusion between vision and text.
\ned{However, our work uses two gating parameters each for the six adapter layers that we add, totaling 12 gating parameters, which is considerably fewer than in their work, which uses two trainable gating matrices of size $2048 \times 512$ and $T \times 512$ where $T$ is the number of input text tokens.}
Furthermore, while the average of the gating parameters used by \citet{wu_good_2021} tended towards 0.0 (consequently weighing vision information lower) as more training is done, we show in this work how the use of vision-based masking allows the training of our gating mechanism to use more of the vision information.

\section{Methods}
\label{sec:methods}

We take a similar approach that \citet{alayrac_flamingo_2022} used to create their generative vision-language model, Flamingo, while adapting their approach for the MMT task.

Flamingo is a generative decoder-only vision-language model created by combining a pre-trained generative language model and a pre-trained vision model, where vision and text interactions are modeled by via gated vision-text cross-attention layers inserted before each decoder layer.
Then, the model is incrementally converted from using only text information to using both vision and text information by freezing the pre-trained portions of the model.
The gating values are set to 0.0 at the beginning of training in order that the vision-language model initially performs equivalently to the language model, and as training progresses the gating values diverge from 0 via back-propagation and consequence learns to use vision information gradually.

Analogously, we start from a pre-trained Transformer-based text-only MT model and a pre-trained vision model to create an MMT model by inserting a vision-text cross-attention layer before each encoder layer. Using trainable gating parameters, we incrementally convert the model from using only text information to using both vision and text information to perform translation. We call our model GRAM (Gating and Residual Adapter-based Model).

While trainable gating parameters have been used in previous work for MMT \citep{wu_good_2021, zhang_neural_2020, lin_dynamic_2020, yin_novel_2020}, our work is unique in the much lower number of gating parameters and in that it allows for the smooth transition of the model from performing as an MT model to performing as an MMT model.

Both the Flamingo model and our model were trained using the next-token prediction task, as is typical for text-only machine translation. Unlike Flamingo, which is a decoder-only model, our model is an encoder-decoder model. \ned{We inserted the vision-text layers before each of the encoder layers only, as we found it to perform better than inserting vision-text layers before the decoder layers only or before both the encoder and decoder layers (Appendix \ref{sec:encdec}).}
Aside from the perceiver resampler module and the gated vision-text cross attention layers used in Flamingo model, which we use to convert our model from an MT model to an MMT model, our \model\ model follows the original text-only Transformer MT model's hyper-parameters, layers, and training objectives as closely as possible.

\subsection{\model\ model architecture}
\label{sec:arch}

\begin{figure}[!t] %
\centering
\includegraphics[width=\linewidth]{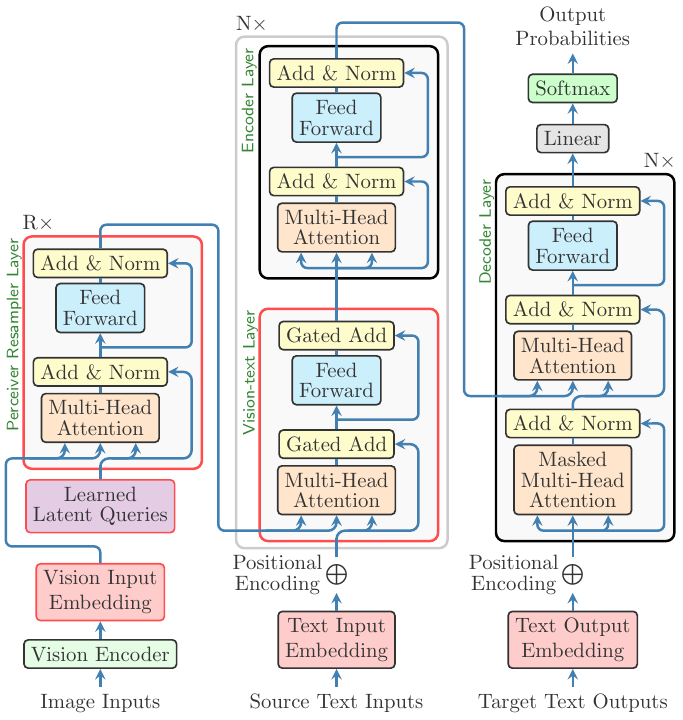}%
\caption{Multimodal translation architecture, where multimodal components are incorporated into the Transformer translation model introduced by \citet{vaswani_attention_2017}. The parameters in the model bordered by red are initialized randomly and updated for training, while the parameters in the pre-trained vision encoder and the pre-trained Transformer translation model bordered by black are frozen. The gating parameters in the vision-text layers are updated using back-propagation, allowing us to smoothly transition from a text-only translation model into a multimodal translation model.
}
\label{fig:model}
\end{figure}

We start with a pre-trained Transformer translation model introduced by \citet{vaswani_attention_2017} and add lightweight multimodal components (Figure \ref{fig:model}). We use a pre-trained vision encoder, CLIP, to encode the input images (Section \ref{sec:visionencoder}). We then link the vision encodings to the Transformer translation model using two components, the perceiver resampler (Section \ref{sec:resampler}) and the vision-text layers (Section \ref{sec:xattn}). The vision encodings, which can come from an arbitrary number of images, are converted into a fixed number of vision tokens using the perceiver resampler. Then, interactions between the vision tokens and the text embeddings are  modeled using the vision-text cross-attention layers. The vision-text layers are incorporated into the Transformer layers by interleaving the vision-text layers and the original self-attention layers of the Transformer encoder.

In more detail, given an input sequence of text tokens $\bm{t} = (t_1, \ldots, t_n)$ and images $\bm{I} =  (I_1, \ldots, I_l)$ where $n$ and $l$ may vary depending on the number of input text tokens and images, the output token sequence is generated auto-regressively as follows.

The vision encoder maps the images $\bm{I}$ into vision encodings $\bm{v} = (v_1,\ldots,v_l)$ where $v_i \in \bm{R}^e$ and $e$ is the size of the image encodings. The vision input embedding layer maps the vision encodings $\bm{v}$ into vision embeddings $\bm{w} = (w_1,\ldots,w_l)$ where $w_i \in \bm{R}^d$ and $d$ is the size of the text and image embeddings. The text input embedding layer maps the text tokens $\bm{t}$ to text embeddings $\bm{x} = (x_1,\ldots,x_n)$ where $x_i \in \bm{R}^d$. The perceiver resampler remaps the variable number of image embeddings to a constant number of vision tokens $\bm{p} = (p_1,\ldots,p_r)$ where $p_i \in \bm{R}^d$, using the $r$ learned latent queries.

Then, the encoder, consisting of a sequence of interleaved vision-text cross-attention layers and encoder layers, maps the text embeddings $\bm{x}$ and vision tokens $\bm{p}$ into a sequence of representations $\bm{z} = (z_1,\ldots,z_n$) where $z_i \in \bm{R}^d$. Given $\bm{z}$, the decoder generates the output probabilities for the next output token in an auto-regressive manner, thus producing the output token sequence, $y_1,\ldots,y_m$.

\subsubsection{Vision encoder}
\label{sec:visionencoder}

We use a pre-trained vision-language model, CLIP \citep{radford_clip_2021}, to encode the input images. CLIP was trained on 400 million image-text pairs using a contrastive image-text approach. The vision encodings produced by CLIP contain rich semantic information relevant to vision-language tasks, and it has been shown to perform well on a wide variety of these tasks. We use the vision encoder in CLIP's best performing ViT-L/14@336px model, which outputs vector encodings of length 768.

\subsubsection{Perceiver resampler}
\label{sec:resampler}

The perceiver resampler, used for the Flamingo model, receives a variable number of vision embeddings and outputs a fixed number of vision tokens. 
This concept was initially used to map a large number of inputs to a fixed number of tokens \citep{jaegle_perceiver_2021} and for object detection, where each of the visual tokens corresponds to an object class \citep{carion_detr_2020}.

Given the vision embeddings $\bm{w}$, let $\bm{\lambda} = (\lambda_1,\ldots,\lambda_r)$ be the learned latent queries, and let $\textbf{MHA}$ and $\textbf{FF}$ be the multi-head attention layer and the feed forward layer, respectively. Then, the first perceiver resampler layer $\textbf{PR}$ is $\textbf{PR}(\bm{\lambda}, \bm{w}) = \bm{\lambda'} + \textbf{FF}(\bm{\lambda'})$ where $\bm{\lambda'} = \bm{\lambda} + \textbf{MHA}(K{=}[\bm{w},\bm{\lambda}], V{=}[\bm{w},\bm{\lambda}], Q{=}\bm{\lambda})$ and $[\bm{w},\bm{\lambda}]$ is the concatenation of the two vectors.
Then, the perceiver resampler layers continue with $\bm{\lambda}{\leftarrow}\textbf{PR}(\bm{\lambda}, \bm{w})$ for $R$ layers. The vision tokens $\bm{p}{\leftarrow}\bm{\lambda}$ are outputted by the final perceiver resampler layer.

\subsubsection{Vision-text layer}
\label{sec:xattn}

Similar to the Flamingo model, in order to smoothly train our MMT model to ensure it behaves at the beginning of training like the pre-trained MT model and behaves at the end of training like an MMT model, we insert vision-text cross-attention layers before each of the original Transformer encoder layers and we use a gating mechanism for each of the vision-text layers.

Given the vision tokens $\bm{p}$ output by the perceiver resampler and the input text embeddings $\bm{x}$, let $g_a$ and $g_f$ be the learnable gating parameters for the multi-head attention layer $\textbf{MHA}$ and the feed forward layer $\textbf{FF}$ respectively, with $\gamma_a = \text{tanh}(g_a)$, $\gamma_f = \text{tanh}(g_f)$. Then, the first gated cross-attention layer $\textbf{GCA}$ is $\textbf{GCA}(\bm{x}, \bm{\lambda}) = \bm{x'} + \gamma_f \, \textbf{FF}(\bm{x'})$ where $\bm{x'} = \bm{x} + \gamma_a \, \textbf{MHA}(K{=}\bm{p}, V{=}\bm{p}, Q{=}\bm{x})$.
The gated cross-attention layers then continue with $\bm{x}{\leftarrow}\textbf{E}(\textbf{GCA}(\bm{x}, \bm{\lambda}))$ for $N$ layers where \textbf{E} is the original Transformer encoder layer following the cross-attention layer.

\ned{Gating parameters are set to 0.0 at the start of training and so it passes the text embeddings $\bm{x}$ through without modification.}
As training continues and as more vision information is used, $\abs{g_a}$ and $\abs{g_f}$ increases via back-propagation; consequently $\abs{\text{tanh}(g_a)}$ and $\abs{\text{tanh}(g_f)}$ approaches 1.0, since the tanh function maps the gating parameters $g_a$ and $g_f$ to be between -1.0 and 1.0

Since the gating parameters initially start at 0.0, vision information is ignored and the model performs as well as the text-only Transformer. During the training process the gating parameters are updated to gradually incorporate vision information for the multimodal translation task. The gating parameters can be seen as a proxy for how much vision information is used by the model.

\subsubsection{Model hyper-parameters}

During training, only the multimodal components are updated, while the vision encoder and the rest of the parameters in the text-only Transformer are kept frozen. For the vision encoder, we use pre-trained weights from the CLIP vision encoder model and ignore CLIP's text encoder model\footnote{See https://github.com/openai/CLIP to download weights.}. For the text-only translation components, we use weights from the pre-trained MT model from FAIR's WMT19 submission\footnote{The weights are from the transformer.wmt19.de-en single model located in the pytorch/fairseq torch hub. See https://github.com/facebookresearch/fairseq for details.}. Since our model uses FAIR's WMT19 MT model, we use the same byte-pair encoding (BPE) and vocabulary used by the MT model. Since the text-only portion of the model is frozen, training is relatively fast, typically 3 batches/sec using two Nvidia V100 GPUs where each batch contains 3,584 tokens.

Since we use the FAIR's WMT19 text-only Transformer as the starting point, we use those hyper-parameters for our additional layers unless otherwise specified.
For the perceiver resampler, we use two layers, i.e., $R = 2$, as was done for Flamingo. For both the perceiver resampler and the vision-text cross attention layers, we use the same parameters as in the text-only Transformer model, except for the number of attention heads being 16 and the intermediate feed-forward layer size being 4,096. The number of parameters are detailed in Appendix \ref{sec:paramnums}).

\subsection{Training}
\label{sec:training}

Beginning with the pre-trained text-only translation model, we add vision embedding layers and gated adapter layers that to the translation model to create a multimodal translation model (Section \ref{sec:arch}). Then, setting the initial gating values to 0.0, which allows our MMT model to perform equivalently to the MT model, we freeze the text-only parameters and train the additional vision-text parameters. We first pre-train the vision-text parameters of our model (Section \ref{sec:pre_train}) and then fine-tune the vision-text parameters using the Multi30k dataset (Section \ref{sec:fine_tune}). During training, the gating value diverge from 0.0 as more vision information gets used by the model. 

\subsubsection{Pre-training}
\label{sec:pre_train}

The intent of the pre-training step is to force the model to use contextually relevant image information by masking vision related words in the source sentence while performing the translation task. We pre-train our model on a dataset collated using vision-based masking of source sentences \ned{that we call the CR dataset}.

\ned{First, we translate 2,878,999 of the English captions in the Conceptual Captions (CC) dataset \citep{sharma_conceptual_2018} that had images available} to German using FAIR's WMT19 translation model, and then perform vision-based masking on the English captions.

For vision-based masking, we create a list of vision related phrases, or topic phrases, by using the VinVL object detector \citep{zhang_vinvl_2021} against the CC images. VinVL is able to detect 1,848 object classes and 524 attribute classes, resulting in a much richer possible vocabulary than other object detectors. With relatively high thresholds of 0.8 for object classes and 0.7 for attribute classes, we create a list of 7,494 ``\texttt{attribute} \texttt{object}'' combinations, such as ``red car''.

Then, for each English-German sentence pair, we search for topic phrases in the English sentence. For each topic phrase we find, we replaced it with the \texttt{<unk>} token (as we are restricted to using tokens present in the pre-trained FAIR WMT19 model, the \texttt{<unk>} token is the closest available token to a mask token). This results in an MMT dataset of 2,663,331 (masked source text, target text, image) triplets.

In addition, we also concatenate \ned{to the CR dataset} 2,878,999 (\texttt{<unk>}, target text, image) triplets created from each of the captions in the CC dataset to further force the usage of vision information to generate text.

Furthermore, so that the model does not overfit to inputs that always contain image information while still maintaining the capacity to translate complex sentences, we concatenate \ned{to the CR dataset} 1,183,301 (source text, target text, $\varnothing$) triplets created from the RAPID 2019 \citep{kocmi_wmt_findings_2022} dataset.

\ned{We train our \model\ model using the typical cross-entropy loss for machine translation. The optimization details for the pre-training step are described in Appendix \ref{sec:optim_pretrain}.}

\begin{table}[t]
\centering
\footnotesize
\begin{tabular}{*{4}{c}}
\toprule
Dataset        & Only text & With image & Total     \\ \midrule %
CR             & 1,183,301     & 5,542,330  & 7,725,631 \\
M30k           & 29,000    & 29,000     & 58,000    \\
\bottomrule
\end{tabular}
\caption{Training datasets used in this work. CR is the augmented Conceptual Captions and RAPID2019 datasets described in Section \ref{sec:pre_train} that we use for pre-training. M30k is the augmented Multi30k dataset used for fine-tuning and is described in Section \ref{sec:fine_tune}. ``Only text'' is the number of examples in the dataset with no associated image. ``With image'' is the number of examples with one or more associated images. ``Total'' is the total number of examples in the dataset.}
\label{tab:datasets}
\end{table}

\subsubsection{Training against Multi30k}
\label{sec:fine_tune}

{\bf Fine-tuning}.
\ned{We use the same vision-based masking described in Section \ref{sec:pre_train} for the source sentences in the Multi30k training set, which resulted in 29,000 masked source text, target text, and image triplets. We refer to this resulting dataset as M30k.
Since the Multi30k dataset contains only 29,000 examples, fine-tuning after the above pre-training step resulted in much better performance compared to directly training against the M30k dataset (Section \ref{sec:experiments}).}

Note that we train our model using a concatenation of the Multi30k training set with images and the Multi30k training set without images. This is to account for evaluation artifacts where the model performance when given both text and image input is higher than model performance with only text input, but the result is only due to the model overfitting on training data that only has (source text, target text, image) triplets and no examples of (source text, target text, $\varnothing$) triplets. \ned{We also explore fine-tuning of three other dataset variations including the original unmasked Multi30k dataset, which we discuss in Appendix \ref{sec:variations}.
The optimization details are described in Appendix \ref{sec:optim_multi30k}.}

{\bf Direct training.} We also directly training using the above described Multi30k dataset without the pre-training step for comparison. Due to its small size, we also explored directly training against the Multi30k dataset using smaller perceiver resampler and vision-text layers, \ned{and found performance to be similar (Appendix \ref{sec:multi30k}).} Thus, we show performance results using the same model sizes.

\begin{table}[t]
\centering
\footnotesize
\begin{tabular}{*{6}{c}}
\toprule

Label & \multicolumn{1}{c}{\scriptsize{CoMMuTE}} & \multicolumn{2}{c}{\scriptsize{Multi30k}} & \multicolumn{2}{c}{\scriptsize{newstest}} \\ \hline

& \multicolumn{1}{c}{} & 2016 & 2017 & 2019 & 2020 \\
\hline
& Score & \multicolumn{4}{c}{BLEU4} \\
\hline
\multicolumn{6}{c}{Multimodal inputs} \\
\modelv{CR}           & 0.57 & 39.2 & 36.8 & \multicolumn{2}{c}{} \\
\modelw{CR}{M30k}     & \bf{0.61} & 46.5 & \bf{43.6} & \multicolumn{2}{c}{} \\
\modelv{M30k}         & 0.50 & 45.9 & 42.7 & \multicolumn{2}{c}{} \\
\scriptsize{Gated Fusion}          & 0.50 & 42.0 & 33.6 & \multicolumn{2}{c}{} \\
VGAMT                 & 0.59 & 43.3 & 38.3 & \multicolumn{2}{c}{} \\
\hline
\multicolumn{6}{c}{Text inputs only} \\
\scriptsize{FAIR-WMT19}    & 0.50 & 40.7 & 37.7 & 40.6 & \bf{36.2} \\
\modelv{CR}            & 0.50 & 40.2 & 37.8 & 40.6 & 35.4 \\
\modelw{CR}{M30k}      & 0.50 & 46.4 & 42.9 & \bf{42.7} & \bf{36.2} \\
\modelv{M30k}          & 0.50 & 45.9 & 42.8 & 36.1 & 26.8 \\
RMMT                   & 0.50 & 41.5 & 33.0 &  1.3 &  0.8 \\
\hline
\multicolumn{6}{c}{Non-matching inputs} \\
\modelv{CR}            & 0.51 & 39.0 & 36.7 & 42.1 & 35.6 \\
\modelw{CR}{M30k}      & 0.51 & \bf{46.6} & 43.2 & 42.0 & \bf{36.2} \\
\modelv{M30k}          & 0.50 & 45.9 & 42.8 & 36.1 & 26.8 \\
\scriptsize{Gated Fusion}           & 0.50 & 42.0 & 33.6 &  1.3 & 0.6 \\
\bottomrule

\end{tabular}
\caption{Performance results for English to German (en-de) translations. The label FAIR-WMT19 shows our model's performance before our training process, i.e., the original text-only Transformer's performance. \modelv{CR} is our model pre-trained on the CR dataset; \modelw{CR}{M30k} is our model pre-trained on CR and fine-tuned on Multi30k; \modelv{M30k} is our model trained on Multi30k without the pre-training step; Gated Fusion and RMMT are our evaluations of the models published by \citet{wu_good_2021}; VGAMT is the reported performance of the model published by \citet{futeral_tackling_2022}. ``Text inputs only'' shows performance of when only the source text is given. ``Multimodal inputs'' shows the performances when both source text and image is used as input.
``Non-matching inputs'' shows performance when source text along with a random image is used as input.
}
\label{tab:main}
\end{table}

\section{Results and Discussion}
\label{sec:experiments}

In previous work \citep{vijayan_multimodal_2023}, we argued that MMT models should be evaluated by measuring 1) their use of visual information to aid in the translation task and 2) their ability to translate complex sentences such as done for text-only machine translation. We use the evaluation framework we proposed in that work here. 
We evaluate model performances against 1) the CoMMuTE \citep{futeral_tackling_2022} test set, 2) the Multi30k \citep{elliott_multi30k_2016} test sets, and 3) \ned{the WMT news translation task \citep{kocmi_wmt_findings_2022} test sets (newstest)} using CoMMuTE score and BLEU4 calculated using SacreBLEU \citep{post_bleu_2018}.

\ned{The main evaluation results are shown in Table \ref{tab:main}.
The label FAIR-WMT19 shows our model's performance before our training process, i.e., the original text-only Transformer's performance.
\modelv{CR} is our \model\ model pre-trained on the CR dataset (Section \ref{sec:pre_train}). \modelw{CR}{M30k} is our model pre-trained on CR and fine-tuned on Multi30k (Section \ref{sec:fine_tune}). \modelv{M30k} is our model trained on Multi30k without the pre-training step (Section \ref{sec:fine_tune}). We compare against the Gated Fusion and RMMT models \citep{wu_good_2021}, which are both trained solely on the Multi30k dataset, as well as the reported performance of VGAMT \citep{futeral_tackling_2022}, which was introduced along with the CoMMuTE test set.}

\subsection{Pre-training using vision-based masking}

Since we begin with a performant MT model, we expect that our model will retain the high text-only performance of the MT model while transforming into an MMT model. In order to ensure this, we followed the work by \citet{alayrac_flamingo_2022}, where they incrementally transformed a language model into a vision-language model which retaining text-only performance, both in terms of the design of our model architecture and our training process (Section \ref{sec:methods}).

Similar to \citet{alayrac_flamingo_2022}, we found that a pre-training step is necessary to successfully transform the model without performance loss.
When we pre-train our model and then fine-tune against the Multi30k dataset, this results in state-of-the-art performance against the Multi30k test sets and CoMMuTE score (Table \ref{tab:main}, label \modelw{CR}{M30k}), as well as little to no degradation of performance against the newstest datasets.

However, when we train against the Multi30k dataset without pre-training, we achieve good performance in the Multi30k test sets but only 0.5 for the CoMMuTE score (Table \ref{tab:main}, \modelv{M30k}), which indicates that image information is not being used by the model, and degraded performance on the newstest datasets (e.g., 36.2 BLEU4 on newstest2020 for the text-only FAIR-WMT19 model compared to 26.8 BLEU4 for \modelv{M30k}).

While our pre-training step does degrade performance slightly on the newstest datasets compared to the original text-only Transformer (e.g., 36.2 BLEU4  on newstest2020 for the text-only FAIR-WMT19 model compared to 35.4 BLEU4 for the \modelv{CR} model), we note that our pre-training process is relatively rudimentary (Section \ref{sec:pre_train}) while FAIR-WMT19 is a model that was fine-tuned specifically for the news translation task using the news commentary dataset \citep{ng_facebook_2019}.
Interestingly, and contrary to expectations, fine-tuning on the Multi30k dataset after pre-training improves performance against the newstest2019 and newstest2020 datasets, which might indicate that the FAIR-WMT19 model is overfitted to the news commentary dataset.

\subsection{Training against Multi30k without pre-training}
\label{sec:non_matching}

Due to the small size of the Multi30k training set, it is expected that models trained against Multi30k without pre-training would perform badly against testing sets such as the newstest datasets. For comparison, in the text-only translation domain, MT models such as FAIR-WMT19 are trained on millions of examples and then evaluated against the newstest dataset.
We evaluated the Gated Fusion and RMMT MMT models, introduced by \citet{wu_good_2021} and trained solely on Multi30k, against the newstest datasets. As expected, there is a drastic drop in performance when the models are evaluated against the newstest datasets (Table \ref{tab:main}).

For the Gated Fusion model, we evaluate by associating random images to the source text and evaluate against the newstest datasets. Since the associated images are not necessarily related to the source text, this can be considered non-matching evaluation. For the RMMT model, which takes as input only the source text, and uses the source text to perform image retrieval for the translation task, we simply use the source text to evaluate against the newstest datasets. As shown in Table \ref{tab:main}, while the models perform well against the Multi30k test sets, they perform very badly against the newstest datasets.

In contrast, since our model uses a performant text-only MT model as the starting point, our model performs well when given non-matching inputs while still having high performance against CoMMuTE and the Multi30k testsets.

\subsection{Text-only translations in Multi30k}
\label{sec:textonly}

One point to note when evaluating against the Multi30k test sets is that most of its captions do not require the image in order to be correctly translated due to the captions being unambiguous. Specifically, \citet{futeral_tackling_2022} analyzed the Multi30k Test2016 and Test2017 and showed that only 2.1\% and 2.0\%, respectively, of the examples in the test sets have ambiguous source sentences that can be resolved using the associated images.
Thus, we expect that correct translations can be achieved with the text alone without the associated images for the vast majority of the remaining examples.
Fitting our expectations, we see that state-of-the-art performance on the Multi30k test sets can be achieved without making use of image information at all (Table \ref{tab:main}, the ``Text inputs only'' rows).

Since high performance can be achieved on the Multi30k test sets without the use of contextual images, it is important that an evaluation framework such as the CoMMuTE evaluation framework that can confirm that visual information is being used to aid in the translation task should always be used in conjunction with the Multi30k test sets when evaluating MMT models.

\begin{figure}[t]
\centering
\scriptsize
\begin{tabular}{*{2}{c}}
\toprule
\multicolumn{2}{c}{Input: Get away from the float!} \\
\includegraphics[height=0.2\linewidth]{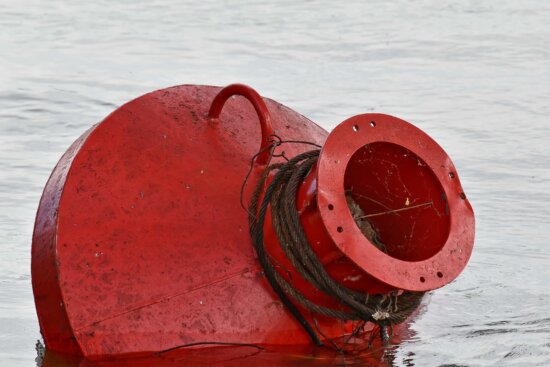} 
 & \includegraphics[height=0.2\linewidth]{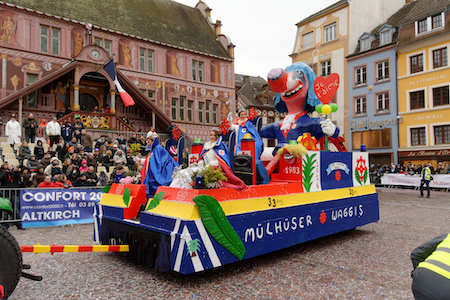} \\ 
Ref: Veg vom Schwimmer! & Output: Veg vom Karnevalswagen! \\
Output: Veg vom Schwimmer! & Output: Veg vom Festwagen! \\
\midrule
\multicolumn{2}{c}{Input: A biker on the road.} \\
\includegraphics[height=0.2\linewidth]{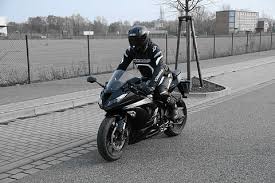} 
 & \includegraphics[height=0.2\linewidth]{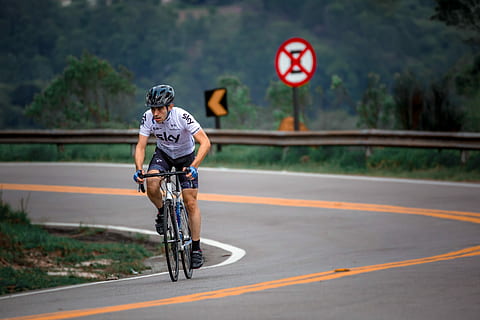} \\ 
Ref: Ein biker auf der Stra{\ss}e. & Ref: Ein Radfahrer auf der Stra{\ss}e. \\
\multirow{2}{14em}{Output: Ein Motorradfahrer auf der Stra{\ss}e.} & \multirow{2}{14em}{Output: Ein Radfahrer auf der Stra{\ss}e.} \\
\\ \bottomrule

\end{tabular}
\caption{Examples from the CoMMuTE test dataset of our model (the \modelw{CR}{M30k} model from Table \ref{tab:main}) resolving ambiguous input text when given contextual images.}
\label{fig:commute}
\end{figure}

\begin{figure}[!ht]
\footnotesize
a) Pre-training.\\
\includegraphics[width=0.98\linewidth]{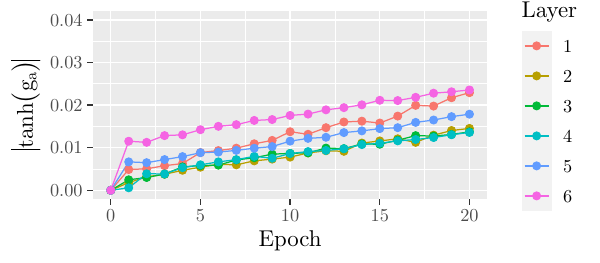}\\
\includegraphics[width=0.98\linewidth]{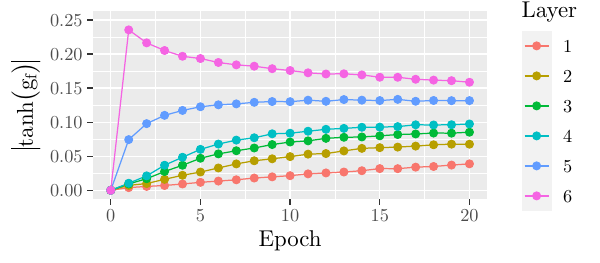}\\
b) Fine-tuning.\\
\includegraphics[width=0.98\linewidth]{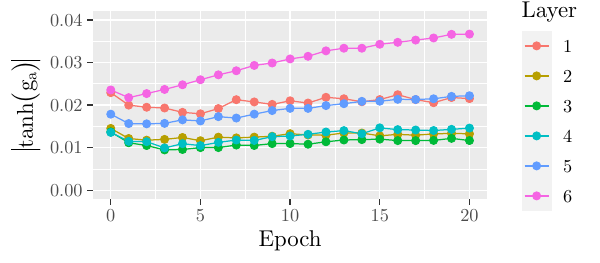}\\
\includegraphics[width=0.98\linewidth]{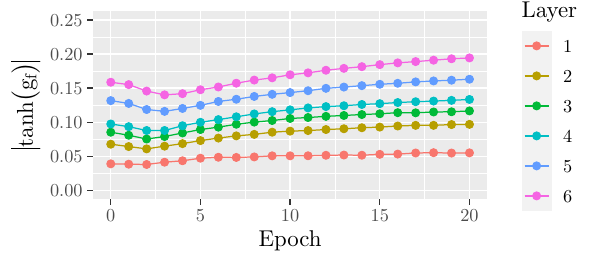}\\
c) Direct training.\\
\includegraphics[width=0.98\linewidth]{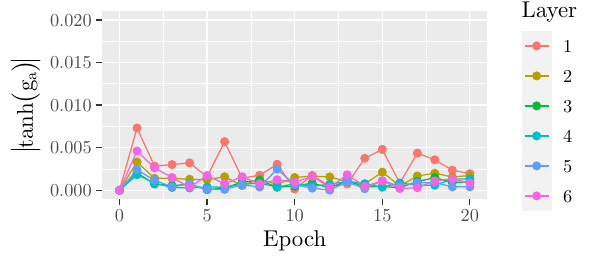}\\
\includegraphics[width=0.98\linewidth]{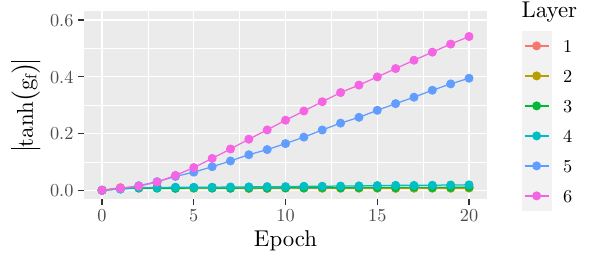}
\caption{Gating values during a) pre-training over the CR dataset, b) fine-tuning over the Multi30k dataset, and c) directly training on the Multi30k dataset. \ned{Layer 1 is the vision-text adapter layer that is closest to the input. Note that some of the gating values overlap in some of the plots.}}
\label{fig:gates}
\end{figure}

\subsection{Gating parameters}
\label{sec:gating}

As in the Flamingo model, our model uses gating parameters to transform from a model that uses only text information to a model that uses both vision and text information to produce outputs.
The gating parameters, explained in Section \ref{sec:arch}, can be viewed as how much the model weighs the image information uses compared to the text information.
Our gating values consists of two sets of values, the attention gating values $g_a$, which directly weigh the incoming image tokens and the input text embeddings, and the feed-forward gating values $g_f$, which weighs the output of the vision-text attention layer and the text embeddings, and can potentially solely use text information in the training set. Thus, the $g_a$ values should be interpreted as the main proxies that indicate how much \ned{image information influences the output of the model.}

Gating parameters have been used previously for MMT, with \citet{wu_good_2021} having explored in detail how gating parameters that weighed vision and text information are affected in MMT models. For their model, as training progressed, the average value gating parameters tended towards 0.0, indicating that their model learned to not use image information as training progressed.

In contrast, our gating parameters did not trend towards 0.0 as training progressed (Figure \ref{fig:gates}), primarily due to the pre-training approach that we employ (as indicated by the difference in the progress of the gating values in pre-training vs. direct training in Figure \ref{fig:gates}).
However, unlike in the Flamingo model, where the maximum of the attention gating values $\abs{\gamma_a} = \abs{\tanh{\gamma_a}}$ reaches around 0.8 towards the end of training, and the maximum of the feed-forward gating values $\abs{\gamma_f} = \abs{\tanh{\gamma_f}}$ reaches 0.95, our gating values reach 0.035 for $\abs{\gamma_a}$ and 0.2 for $\abs{\gamma_f}$.
This suggests that image information is not necessarily as important for the multimodal translation task compared to the Flamingo model, which can perform a wide variety of tasks including visual question answering.

\section{Conclusion}
\label{sec:conclusion}

Text-only machine translation is a much larger domain than multimodal machine translation and many strong models have been developed in the field.
The approach of transforming a language model into a vision-language model was successful demonstrated via Flamingo, and thus have a high probability of working well in the similar task of machine translation.
Following this idea, we designed an MMT model that began as a performant text-only MT model and incrementally transformed it into a MMT model by 1) pre-training using informed vision-based masking of the source text and 2) fine-tuning on Multi30k. We achieved a state-of-the-art performance on the Multi30k 2016 test set of 46.5 BLEU4 score via this approach while retaining high performance against CoMMuTE and the newstest datasets. 
There are many approaches for improving our model including the training process, where the pre-training dataset can be improved using more text-only datasets or augmenting text-only datasets using image retrieval, and model architecture, where techniques such as VMLM can be used to further enforce the use of image information in the model.

\thanksnostar{
Opinions, interpretations, conclusions, and recommendations are those of the authors and are not necessarily endorsed by the United States Government. Cleared for public release on 13 Feb 2024. Originator reference number RH-24-125355. Case number AFRL-2024-0832.
}

\section*{Acknowledgements}

This work is sponsored by the Air Force Research Laboratory
under Air Force contract FA8650-20-D-6207.

\bibliography{zot}

\clearpage
\newpage

\appendix

\section{Number of parameters in the \model\ model}
\label{sec:paramnums} 

The number of parameters in the original text-only Transformer is 269,746,176. While there are also 304,293,888 parameters in the ViT-L/14@336px CLIP vision encoder model that we use, the vision encoder is not used during training since we cache the image encodings to file. We add 68,051,980 parameters via the perceiver resampler and the six vision-text layers, which are that parameters that we optimize over. Thus, the entire model contains 337,798,156 parameters. If we include the vision encoder as well, then the entire model contains 642,092,044 parameters.

\section{Optimization details}
\label{sec:optim}

\subsection{Pre-training}
\label{sec:optim_pretrain}

We use the same optimization hyper-parameters as FAIR's WMT19 model \citep{ng_facebook_2019} with Fairseq \citep{ott_fairseq_2019} as the training and evaluation framework.
For pre-training, we use the Adam optimizer with $\beta_1 = 0.9, \beta_2 = 0.98$, with a warm-up phase of 4,000 steps where we linearly increase the learning rate from $10^{-7}$ to $0.0007$. Each training batch contains 3,584 source/target tokens and we train for 20 epochs. We use the checkpoint from the last epoch for fine-tuning.

\subsection{Training against Multi30k}
\label{sec:optim_multi30k}

{\bf Fine-tuning.}
When we perform fine-tuning, we lower the learning rate to $0.0002$ and train for 20 epochs. Since the Multi30k dataset is small, we use a warm-up phase of 240 steps where we linearly increase the learning rate from $10^{-7}$ to $0.0002$. We select the checkpoint that performs best against the Multi30k validation set with respect to BLEU4 score.

{\bf Direct training.}
When we directly train, we set the learning rate to $0.0007$ and train for 20 epochs using a warm-up phase of 240 steps.

\clearpage
\newpage

\section{Model variations}

\subsection{Where to insert the vision-text adapter layers}
\label{sec:encdec}

\begin{table*}[t]
\centering
\footnotesize
\begin{tabular}{*{11}{c}}
\toprule

Label & PT & FT & \multicolumn{2}{c}{CoMMuTE} & \multicolumn{4}{c}{Multi30k} & \multicolumn{2}{c}{newstest} \\ \midrule

& \multicolumn{2}{c}{} & \multicolumn{2}{c}{} & 2016 & 2017 & coco & 2018 & 2019 & 2020 \\ \midrule

& \multicolumn{2}{c}{} & Score & \multicolumn{7}{c}{BLEU4} \\
\midrule
\multicolumn{11}{c}{Multimodal inputs} \\

\modelv{CR}                 & CR &                & 0.57 & 35.08 & 39.17 & 36.79 & 31.45 & 35.72 & \multicolumn{2}{c}{} \\
\modelvdeconly{CR}          & CR &                & 0.55 & 32.59 & 41.16 & 37.54 & 33.46 & 36.64 & \multicolumn{2}{c}{} \\
\modelvencdec{CR}           & CR &                & 0.52 & 34.14 & 39.56 & 37.45 & 31.34 & 35.94 & \multicolumn{2}{c}{} \\
\midrule
\multicolumn{11}{c}{Text inputs only} \\
\scriptsize{FAIR-WMT19}                &    &     & 0.50 & 32.63 & 40.66 & 37.70 & 33.97 & 36.45 & 40.62 & 36.20 \\
\modelv{CR}                 & CR &                & 0.50 & 31.98 & 40.22 & 37.75 & 32.81 & 36.41 & 40.56 & 35.35 \\
\modelvdeconly{CR}          & CR &                & 0.50 & 30.01 & 40.85 & 37.19 & 33.36 & 35.84 & 38.36 & 33.79 \\
\modelvencdec{CR}           & CR &                & 0.50 & 30.61 & 40.03 & 37.80 & 32.34 & 36.11 & 40.18 & 34.15 \\
\midrule
\multicolumn{11}{c}{Non-matching inputs} \\
\modelv{CR}                 & CR &                & 0.51 & 30.37 & 39.01 & 36.73 & 32.10 & 35.35 & 42.09 & 35.62 \\
\modelvdeconly{CR}          & CR &                & 0.50 & 33.07 & 41.02 & 37.72 & 33.54 & 36.59 & 42.17 & 36.20 \\
\modelvencdec{CR}           & CR &                & 0.50 & 34.02 & 39.67 & 37.44 & 31.19 & 35.74 & 40.84 & 34.95 \\
\bottomrule

\end{tabular}
\caption{Performance results of our model under various pre-training and fine-tuning conditions for English to German (en-de) translations. The label FAIR-WMT19 shows our model's performance before our training process, i.e., the original text-only Transformer's performance. \modelv{CR} is our model pre-trained on the CR dataset. \modelvdeconly{CR} and \modelvencdec{CR} are variations where the vision-text layers are inserted before the decoder layers only (\modelvdeconly{CR}) and inserted before both the encoder and decoder layers (\modelvencdec{CR}), while the \modelv{CR} model is the variation where the vision-text layers are inserted before the encoder layers only. ``Text inputs only'' shows the performances of our model when only the source text is given and a zero vector is given as the vision encoding. ``Multimodal inputs'' shows the performances of our model when both source text and image is used as input. The model is evaluated against the CoMMuTE \citep{futeral_tackling_2022} testing set, the Multi30k \citep{elliott_multi30k_2016} sets, and the newstest \citep{kocmi_wmt_findings_2022} testing sets using BLEU4, calculated using SacreBLEU \citep{post_bleu_2018}. Both CoMMuTE score and BLEU4 scores against the CoMMuTE test dataset are shown for completeness; since the CoMMuTE sentences are very short, the BLEU4 score for CoMMuTE should be weighed lightly. PT indicates pre-training and FT indicates fine-tuning. The datasets used for pre-training and fine-tuning are described in Table \ref{tab:datasets}.}
\label{tab:encdec}
\end{table*}

For the \model\ model, vision-text cross-attention adapter layers can be added before each of the layers in the Transformer model. Since we modify an encoder-decoder Transformer in order to transform it from an MT model to an MMT model, there are three options for where we add the vision-text layers. One is to insert the vision-text layers before each layer in the Transformer encoder (\modelv{CR}). Second is to insert the vision-text layers before each layer in the Transformer decoder (\modelvdeconly{CR}). Third is to insert the vision-text layers before each layer in both the Transformer encoder and decoder (\modelvencdec{CR}). 

We compare the performance of the three options, the results which are in Table \ref{tab:encdec}. We selected the \modelv{CR} for fine-tuning since the CoMMuTE score was 0.57 compared to CoMMuTE score of 0.55 for \modelvdeconly{CR} and 0.52 for \modelvencdec{CR}.

\subsection{Smaller model variations when directly training against Multi30k}
\label{sec:multi30k}

\begin{table*}[t]
\centering
\footnotesize
\begin{tabular}{*{11}{c}}
\toprule

Label & PT & FT & \multicolumn{2}{c}{CoMMuTE} & \multicolumn{4}{c}{Multi30k} & \multicolumn{2}{c}{newstest} \\ \midrule

& \multicolumn{2}{c}{} & \multicolumn{2}{c}{} & 2016 & 2017 & coco & 2018 & 2019 & 2020 \\ \midrule

& \multicolumn{2}{c}{} & Score & \multicolumn{7}{c}{BLEU4} \\
\midrule
\multicolumn{11}{c}{Multimodal inputs} \\
\modelv{M30k$_o$}           & M30k$_o$ &          & 0.50 & 31.99 & 45.52 & 42.20 & 37.51 & 39.30 & \multicolumn{2}{c}{} \\
\modelv{M30k}               & M30k &              & 0.50 & 27.12 & 45.93 & 42.76 & 37.64 & 38.82 & \multicolumn{2}{c}{} \\
\modelvsmall{M30k$_o$}      & M30k$_o$ &          & 0.50 & 33.61 & 46.41 & 42.29 & 37.83 & 39.71 & \multicolumn{2}{c}{} \\
\modelvsmall{M30k}          & M30k &              & 0.50 & 29.70 & 46.09 & 41.61 & 38.58 & 38.98 & \multicolumn{2}{c}{} \\
\modelvtiny{M30k$_o$}       & M30k$_o$ &          & 0.50 & 33.06 & 46.74 & 42.44 & 38.06 & 39.32 & \multicolumn{2}{c}{} \\
\modelvtiny{M30k}           & M30k &              & 0.50 & 27.12 & 45.93 & 42.76 & 37.64 & 38.82 & \multicolumn{2}{c}{} \\
\midrule
\multicolumn{11}{c}{Text inputs only} \\
\modelv{M30k$_o$}           & M30k$_o$ &          & 0.50 & 31.99 & 45.52 & 42.20 & 37.51 & 39.30 & 37.77 & 28.30 \\
\modelv{M30k}               & M30k &              & 0.50 & 27.12 & 45.93 & 42.76 & 37.64 & 38.82 & 36.09 & 26.81 \\
\modelvsmall{M30k$_o$}      & M30k$_o$ &          & 0.50 & 33.61 & 46.41 & 42.29 & 37.83 & 39.71 & 37.75 & 27.59 \\
\modelvsmall{M30k}          & M30k &              & 0.50 & 29.70 & 46.09 & 41.61 & 38.58 & 38.98 & 36.71 & 27.89 \\
\modelvtiny{M30k$_o$}       & M30k$_o$ &          & 0.50 & 33.06 & 46.74 & 42.44 & 38.06 & 39.32 & 37.09 & 28.12 \\
\modelvtiny{M30k}           & M30k &              & 0.50 & 29.38 & 46.21 & 42.20 & 38.08 & 38.88 & 37.37 & 28.21 \\
\midrule
\multicolumn{11}{c}{Non-matching inputs} \\
\modelv{M30k$_o$}           & M30k$_o$ &          & 0.50 & 31.99 & 45.52 & 42.20 & 37.51 & 39.30 & 37.77 & 28.30 \\
\modelv{M30k}               & M30k &              & 0.50 & 27.12 & 45.93 & 42.76 & 37.64 & 38.82 & 36.09 & 26.81 \\
\modelvsmall{M30k$_o$}      & M30k$_o$ &          & 0.50 & 33.61 & 46.41 & 42.29 & 37.83 & 39.71 & 37.75 & 27.59 \\
\modelvsmall{M30k}          & M30k &              & 0.50 & 29.70 & 46.09 & 41.61 & 38.58 & 38.98 & 36.71 & 27.89  \\
\modelvtiny{M30k$_o$}       & M30k$_o$ &          & 0.50 & 33.06 & 46.74 & 42.44 & 38.06 & 39.32 & 37.09 & 28.12 \\
\modelvtiny{M30k}           & M30k &              & 0.50 & 29.38 & 46.21 & 42.20 & 38.08 & 38.88 & 37.37 & 28.21 \\
\bottomrule

\end{tabular}
\caption{Performance results of our model under various pre-training and fine-tuning conditions for English to German (en-de) translations. The label FAIR-WMT19 shows our model's performance before our training process, i.e., the original text-only Transformer's performance. \modelv{CR} is our model pre-trained on the CR dataset; \modelw{CR}{M30k} is our model pre-trained on CR and fine-tuned on Multi30k; \modelv{M30k} is our model trained on Multi30k without the pre-training step. \modelvsmall{M30k} and \modelvtiny{M30k} are smaller variations of the \modelv{M30k} model. The datasets used for pre-training and fine-tuning are described in Table \ref{tab:datasets}.}
\label{tab:multi30k}
\end{table*}

We also explored smaller models when directly training against Multi30k due to the small size of the dataset. For the first smaller model, we set the number of attention heads to 8 and intermediate feed-forward layer size to 2,048 of the vision-text cross-attention layers (\modelvsmall{M30k$_o$} and \modelvsmall{M30k}). For the second smaller model, we set the number of attention heads to 4 and intermediate feed-forward layer size to 1,024 of the vision-text cross-attention layers (\modelvtiny{M30k$_o$} and \modelvtiny{M30k}). As shown in Table \ref{tab:multi30k}, we found performance to be similar.

\clearpage
\newpage

\section{Dataset variations}
\label{sec:variations}

\begin{table}[b]
\centering
\footnotesize
\begin{tabular}{*{4}{c}}
\toprule
Dataset        & Only text & With image & Total \\ \midrule
CR             & 1,183,301  & 5,542,330  & 7,725,631 \\
M30k$_o$           & 29,000     & 29,000     & 58,000 \\
M30k       & 29,000     & 29,000     & 58,000 \\
M30k$_o$/ncv14     & 29,000     & 339,099    & 367,099 \\
M30k/ncv14 & 29,000     & 338,099    & 367,099 \\
\bottomrule

\end{tabular}
\caption{Training datasets used in this work. CR is the augmented Conceptual Captions dataset described in Section \ref{sec:pre_train} that we use for pre-training. The remaining datasets are used for fine-tuning and is described in Section \ref{sec:fine_tune}. ``Only text'' is the number of examples in the dataset with no associated image. ``With image'' is the number of examples with one or more associated images. ``Total'' is the total number of examples in the dataset.}
\label{tab:variations}
\end{table}

\begin{table*}[t]
\centering
\footnotesize
\begin{tabular}{*{11}{c}}
\toprule

Label & PT & FT & \multicolumn{2}{c}{CoMMuTE} & \multicolumn{4}{c}{Multi30k} & \multicolumn{2}{c}{newstest} \\ \midrule

& \multicolumn{2}{c}{} & \multicolumn{2}{c}{} & 2016 & 2017 & coco & 2018 & 2019 & 2020 \\ \midrule

& \multicolumn{2}{c}{} & Score & \multicolumn{7}{c}{BLEU4} \\
\midrule
\multicolumn{11}{c}{Multimodal inputs} \\

\modelv{CR}                 & CR &                & 0.57 & 35.08 & 39.17 & 36.79 & 31.45 & 35.72 & \multicolumn{2}{c}{} \\
\modelw{CR}{M30k$_o$}       & CR & M30k$_o$       & 0.58 & 33.03 & 47.11 & 43.75 & 39.48 & 40.94 & \multicolumn{2}{c}{} \\
\modelw{CR}{M30k}           & CR & M30k           & 0.61 & 35.03 & 46.50 & 43.57 & 39.10 & 40.40 & \multicolumn{2}{c}{} \\ 
\modelw{CR}{M30k$_o$/ncv14} & CR & M30k$_o$/ncv14 & 0.58 & 33.99 & 47.38 & 42.95 & 39.83 & 40.92 & \multicolumn{2}{c}{} \\
\modelw{CR}{M30k/ncv14}     & CR & M30k/ncv14     & 0.63 & 34.88 & 46.57 & 43.58 & 39.78 & 41.03 & \multicolumn{2}{c}{} \\
\modelv{M30k$_o$}           & M30k$_o$ &          & 0.50 & 31.99 & 45.52 & 42.20 & 37.51 & 39.30 & \multicolumn{2}{c}{} \\
\modelv{M30k}               & M30k &              & 0.50 & 27.12 & 45.93 & 42.76 & 37.64 & 38.82 & \multicolumn{2}{c}{} \\
\midrule
\multicolumn{11}{c}{Text inputs only} \\
\scriptsize{FAIR-WMT19}                &    &     & 0.50 & 32.63 & 40.66 & 37.70 & 33.97 & 36.45 & 40.62 & 36.20 \\
\modelv{CR}                 & CR &                & 0.50 & 31.98 & 40.22 & 37.75 & 32.81 & 36.41 & 40.56 & 35.35 \\
\modelw{CR}{M30k$_o$}       & CR & M30k$_o$       & 0.50 & 31.25 & 47.10 & 43.08 & 38.48 & 40.82 & 42.64 & 36.00\\
\modelw{CR}{M30k}           & CR & M30k           & 0.50 & 32.11 & 46.43 & 42.88 & 37.88 & 40.35 & 42.66 & 36.22 \\ 
\modelw{CR}{M30k$_o$/ncv14} & CR & M30k$_o$/ncv14 & 0.50 & 31.17 & 47.40 & 43.30 & 38.86 & 40.70 & 41.80 & 36.44 \\
\modelw{CR}{M30k/ncv14}     & CR & M30k/ncv14     & 0.50 & 32.95 & 46.65 & 43.06 & 38.95 & 40.73 & 41.86 & 36.46 \\
\modelv{M30k$_o$}           & M30k$_o$ &          & 0.50 & 31.99 & 45.52 & 42.20 & 37.51 & 39.30 & 37.77 & 28.30 \\
\modelv{M30k}               & M30k &              & 0.50 & 27.12 & 45.93 & 42.76 & 37.64 & 38.82 & 36.09 & 26.81 \\
\midrule
\multicolumn{11}{c}{Non-matching inputs} \\
\modelv{CR}                 & CR &                & 0.51 & 30.37 & 39.01 & 36.73 & 32.10 & 35.35 & 42.09 & 35.62 \\
\modelw{CR}{M30k$_o$}       & CR & M30k$_o$       & 0.52 & 32.17 & 47.08 & 42.97 & 38.55 & 41.12 & 42.31 & 36.12 \\
\modelw{CR}{M30k}           & CR & M30k           & 0.51 & 31.22 & 46.56 & 43.19 & 37.94 & 40.75 & 42.04 & 36.18 \\
\modelw{CR}{M30k$_o$/ncv14} & CR & M30k$_o$/ncv14 & 0.50 & 29.39 & 47.24 & 43.44 & 39.48 & 41.11 & 41.82 & 36.52 \\
\modelw{CR}{M30k/ncv14}     & CR & M30k/ncv14     & 0.51 & 31.69 & 46.37 & 43.06 & 38.90 & 40.72 & 41.78 & 36.27 \\
\modelv{M30k$_o$}           & M30k$_o$ &          & 0.50 & 31.99 & 45.52 & 42.20 & 37.51 & 39.30 & 37.77 & 28.30 \\
\modelv{M30k}               & M30k &              & 0.50 & 27.12 & 45.93 & 42.76 & 37.64 & 38.82 & 36.09 & 26.81 \\
\bottomrule

\end{tabular}
\caption{Performance results of our model under various pre-training and fine-tuning conditions for English to German (en-de) translations. The label FAIR-WMT19 shows our model's performance before our training process, i.e., the original text-only Transformer's performance. \modelv{CR} is our model pre-trained on the CR dataset; \modelw{CR}{M30k} is our model pre-trained on CR and fine-tuned on Multi30k; \modelv{M30k} is our model trained on Multi30k without the pre-training step. The datasets used for pre-training and fine-tuning are described in Table \ref{tab:datasets}.}
\label{tab:variations}
\end{table*}

We explore four variations of our model where we fine-tune against four datasets: M30k$_o$, M30k, M30k$_o$/ncv14, and M30k/ncv14 (Table \ref{tab:variations}). The results are shown in Table \ref{tab:variations}.

M30k$_o$ is the original Multi30k dataset. However, we train our model using a concatenation of the Multi30k training set with images and the Multi30k training set without images. This is to account for evaluation artifacts where the model performance when given both text and image input is higher than model performance with only text input, but the result is only due to the model overfitting on training data that only has (source text, target text, image) triplets and no examples of (source text, target text, $\varnothing$) triplets.

M30k is the Multi30k dataset with vision-based masking of the source sentences as done in Section \ref{sec:pre_train}. For each (source text, target text, image), we search for topic phrases (see Section \ref{sec:pre_train}) in the source sentence and replace each instance of the topic phrase with the \texttt{<unk>} token. We also concatenate the original Multi30k dataset with the (source text, target text, image(s)) triplets and the Multi30k dataset with images removed (source text, target text, $\varnothing$) to this.

M30k$_o$/ncv14 and M30k/ncv14 are the concatenation of M30k$_o$ and M30k, respectively, to the news commentary v14 dataset. The news commentary v14, a news translation dataset comprising X sentence pairs, has been used by \citet{ng_facebook_2019} in their fine-tuning step in order to perform well against the newstest testing sets.

{\bf Optimization details for the dataset variants.} When we perform fine-tuning, we lower the learning rate to $0.0002$ and train for 20 epochs. Since the Multi30k dataset is small, for M30k$_o$ and M30k we use a warm-up phase of 240 steps where we linearly increase the learning rate from $10^{-7}$ to $0.0002$. We select the checkpoint that performs best against the Multi30k validation set with respect to BLEU4 score.
For M30k$_o$/ncv14 and M30k/ncv14, we use a warm-up phase of 1200 steps where we linearly increase the learning rate from $10^{-7}$ to $0.0002$. We create a validation set from the concatenation of the WMT19 validation set and the Multi30k validation set and select the checkpoint that performs best against the validation set with respect to BLEU4 score.

\subsection{Simultaneously fine-tuning Multi30k and a text-only dataset}

Since the pre-training step does degrade performance on the newstest datasets (e.g., 36.2 BLEU4  on newstest2020 for the text-only FAIR-WMT19 model compared to 35.4 BLEU4 for the \modelv{CR} model), and fine-tuning against Multi30k alone only slightly improves this performance, we explore how to fine-tune our model such that we preserve the performance on the Multi30k test sets and improve the performance on the newstest datasets.

\citet{ng_facebook_2019} used the news commentary dataset \citep{kocmi_wmt_findings_2022}, a news translation dataset, as the final fine-tuning step in order to improve performance against the newstest datasets. Similarly, we perform fine-tuning on a concatenation of the Multi30k and news commentary v14 dataset, which resulted in improvements in both the newstest datasets and the Multi30k test sets (e.g., 35.4 BLEU4  on newstest2020 for the \modelv{CR} model compared to 36.2 BLEU4 for the \modelw{CR}{M30k/ncv14} model).

\subsection{Fine-tuning without vision-based masking of source text}

Since most of the captions in Multi30k do not require the image in order to be correctly translated due to the captions being unambiguous \citep{futeral_tackling_2022}, MMT models tend to ignore visual information during the training process \citep{caglayan_probing_2019, wu_good_2021}. We are able to quantitatively see this when directly training against the original Multi30k dataset (for \modelv{M30k$_o$}, the CoMMuTE score is 0.5).

So we ask ourselves how we may preserve CoMMuTE performance along with newstest and Multi30k test performances. Since vision-based masking of source sentences was used to improve performance during the pre-training stage, we explore whether it can improve performance during the fine-tuning stage as well.

Thus, we create the M30k and the M30k/ncv14 datasets as described above. The M30k contains masked source sentences from the Multi30k dataset and the M30k/ncv14 dataset is a concatenation of the M30k and the text-only news commentary v14 datasets.
We see that fine-tuning using these datasets preserve the CoMMuTE score much better than when not using informed masking (Table \ref{tab:variations}) while only slightly decreasing BLEU4 scores.

\end{document}